\documentclass[conference]{IEEEtran}
\IEEEoverridecommandlockouts
\usepackage{cite}
\usepackage{amsmath,amssymb,amsfonts}
\usepackage{algorithmic}
\usepackage{graphicx}
\usepackage{textcomp}
\usepackage{xcolor}

\usepackage[normalem]{ulem} 
\useunder{\uline}{\ul}{}
\usepackage{booktabs}
\usepackage{hyperref}
\usepackage{dsfont} 

\def\BibTeX{{\rm B\kern-.05em{\sc i\kern-.025em b}\kern-.08em
    T\kern-.1667em\lower.7ex\hbox{E}\kern-.125emX}}

\usepackage{fancyhdr,lipsum}
\fancypagestyle{firstpage}{
  \fancyhead[C]{\small To appear at the $16^{th}$ IEEE International Conference on Software Testing, Verification and Validation (ICST 2023), Dublin, Ireland.}
  \fancyfoot[C]{\thepage}
}
\begin{document}

\title{Poster: Link between Bias, Node Sensitivity and Long-Tail Distribution in trained DNNs
\thanks{This work was partially supported by Doctoral College Resilient Embedded Systems which is run jointly by TU Wien's Faculty of Informatics and FH-Technikum Wien, and partially by Moore4Medical project funded by the ECSEL Joint Undertaking under grant number H2020-ECSEL-2019-IA-876190.}
}

\author{\IEEEauthorblockN{Mahum Naseer}
\IEEEauthorblockA{\textit{Institute of Computer Engineering} \\
\textit{Technische Universit\"at Wien (TU Wien)}\\
Vienna, Austria \\
\url{https://orcid.org/0000-0003-3096-9536}}
\and
\IEEEauthorblockN{Muhammad Shafique}
\IEEEauthorblockA{\textit{Division of Engineering} \\
\textit{New York University Abu Dhabi (NYUAD)}\\
Abu Dhabi, United Arab Emirates \\
\url{https://orcid.org/0000-0002-2607-8135}}
}

\maketitle
\thispagestyle{firstpage}
\begin{abstract}
Owing to their remarkable learning (and relearning) capabilities, deep neural networks (DNNs) find use in numerous real-world applications. However, the learning of these data-driven machine learning models is generally as good as the data available to them for training. Hence, training datasets with long-tail distribution pose a challenge for DNNs, since the DNNs trained on them may provide a varying degree of classification performance across different output classes. While the overall bias of such networks is already highlighted in existing works, this work identifies the node bias that leads to a varying sensitivity of the nodes for different output classes. To the best of our knowledge, this is the first work highlighting this unique challenge in DNNs, discussing its probable causes, and providing open challenges for this new research direction. We support our reasoning using an empirical case study of the networks trained on a real-world dataset. 
\end{abstract}


\section{Introduction}
The reliance of real-world applications on smart systems based on deep neural networks (DNNs) has been on a constant rise for several years. 
These include the applications in safety-critical systems like autonomous driving and healthcare \cite{health,auto_drive}. 
This raises concerns regarding the reliable and acceptable performance of DNNs for a diverse range of input scenarios, including the ones pertaining to consistent performance of the DNNs trained on long-tail distribution, i.e., the training datasets with a significant portion of total inputs belonging to the \textit{head} class(es) and only a small subset of inputs belonging to the \textit{tail} class(es) \cite{2022longtail}. 

The concerns surrounding long-tail distribution are not ill-founded. 
Numerous available datasets, in fact, comprise of long-tail distribution. 
The MIT-BIH Arrhythmia dataset \cite{arythmia-dataset} contains a significant proportion of normal ECG samples (as opposed to ECG samples indicating arrhythmia).
The IMDB-WIKI dataset \cite{imdb-dataset} comprises of a significant proportion of Caucasian faces.
Wafer map training dataset \cite{wu2014wafer} comprises of a proportion of fault-free wafers (as opposed to faulty wafers). 
Such discrepancy in the number of inputs across different output classes is not always surprising, since the tail classes often present rare events of the real-world. 

It is not also surprising that the DNNs trained on long-tail distribution learn the patterns in head classes better than those in the tail classes due to the availability of ample input samples. 
It has also been observed that such networks also delineate a robustness bias under the influence of noise, i.e., the network is likely to correctly classify even noisy inputs from the head class(es), while the robustness of the tail class(es) against noise is only negligible. 

Orthogonally, the sensitivity of input nodes has also been found to vary \cite{bhatti2022formal}. 
While this variation comes in handy while determining the relevant input nodes for the designated task of the trained DNN \cite{input-sens1,chen2020sensitivity}, it may also pose itself as a concern for applications where a revelation of the sensitive attributes (nodes) may lead to a privacy infringement \cite{sens4fair1_dnn,sens4fair2_dnn}.

However, there is another aspect of concern for DNNs, which is inadvertently linked to those indicated  above - i.e., the (robustness) bias of the individual input nodes - which remains unrecognized in the existing literature. 
This work deals with such node bias, indicating its stealthy existence and the non-triviality of understanding its causes. 
To summarize, the novel contributions of this work are as follows: 
\begin{enumerate}
    \item Defining the concept of node (robustness) bias.
    \item Highlighting the link between robustness bias, node sensitivity, and node bias.
    \item Identifying the existence of node bias and empirically analyzing it in a network trained on a real-world Leukemia dataset.
    \item Discussing the severity of node bias with respect to long-tail distribution of the training dataset.
    \item Elucidating the open challenges pertaining to node bias, in trained DNNs.
\end{enumerate}

\section{Preliminaries} \label{sec:pre}

This section describes the terminologies and concepts used throughout the rest of this paper. \\

\noindent\textbf{Robustness.} \textit{Given a network $\mathcal{N} : \mathcal{X} \rightarrow L(\mathcal{X})$, $\mathcal{N}$ is said to be robust iff the addition of noise $\epsilon \leq N$ to any input $x \in \mathcal{X}$ does not change the output classification of $x$, i.e., $\mathcal{N}(x) = \mathcal{N}(x + \epsilon)$.}

However, given the large (and often infinite) size of the input domain $\mathcal{X}$, it is often infeasible to check the global robustness of the network. Hence, the local robustness of the input domain, surrounding seed inputs, i.e., $x+\epsilon: \epsilon\leq N$ is instead the focus of the practical analysis. \\

\noindent\textbf{Robustness Bias. } \textit{Given a network $\mathcal{N} : \mathcal{X} \rightarrow L(\mathcal{X})$, where $L(\mathcal{X})$ comprises of $\mathcal{C}$ output classes (i.e., $L(\mathcal{X})=\{1,...,\mathcal{C}\}$), robustness bias defines the robustness of individual output classes. This means, robustness bias holds for $\mathcal{N}$ iff the probability of correct classification for inputs $x_k$ belonging to all output classes $k \in L(\mathcal{X})$, under the incidence of noise $\epsilon \leq N$, is equal, i.e., $\forall k \in L(\mathcal{X}), \mathds{P}[\mathcal{N}(x_k) = \mathcal{N}(x_k + \epsilon)] = const$.}

A lack of robustness bias could be attributed to long-tail distribution of the training dataset \cite{bhatti2022formal}, with the probability of correct classification for head classes being higher than that for the tail classes, in the trained DNN.\\

\noindent\textbf{Node Sensitivity. } \textit{Given a network $\mathcal{N}: \mathcal{X} \rightarrow L(\mathcal{X})$, where each input comprises of $n$ input nodes, node sensitivity determines the robustness of individual input nodes under the incidence of the node noise $\eta \leq N$.}

In principle, an input node may be sensitive or insensitive to a specific kind of noise, for instance to the positive noise or the noise bounded by specific constraints. \\

\noindent\textbf{Node (Robustness) Bias. } \textit{Given a network $\mathcal{N} : \mathcal{X} \rightarrow L(\mathcal{X})$, where $L(\mathcal{X})$ comprises of $\mathcal{C}$ output classes (i.e., $L(\mathcal{X})=\{1,...,\mathcal{C}\}$) and each input comprises of $n$ input nodes, node (robustness) bias defines the robustness of individual input nodes for each output class. This means, node (robustness) bias holds for the input node $x_k^i \in x_k$ iff the probability of correct classification for input $x_k$ belonging to all output classes $k \in L(\mathcal{X})$ when noise $\eta \leq N$ is incident to node $x_k^i$, is equal, i.e., $ \forall k \in L(\mathcal{X}), \forall i \in n. \mathds{P}[\mathcal{N}(x_k) = \mathcal{N}(x_k\setminus x_k^i,x_k^i + \eta)] = const$.}

The intuition behind the analysis of node (robustness) bias is to ensure that each input node has a consistent sensitivity for inputs belonging to all output classes.

\section{Proposed Framework}

\begin{figure}[ht]
\begin{center}
\includegraphics[width=\linewidth]{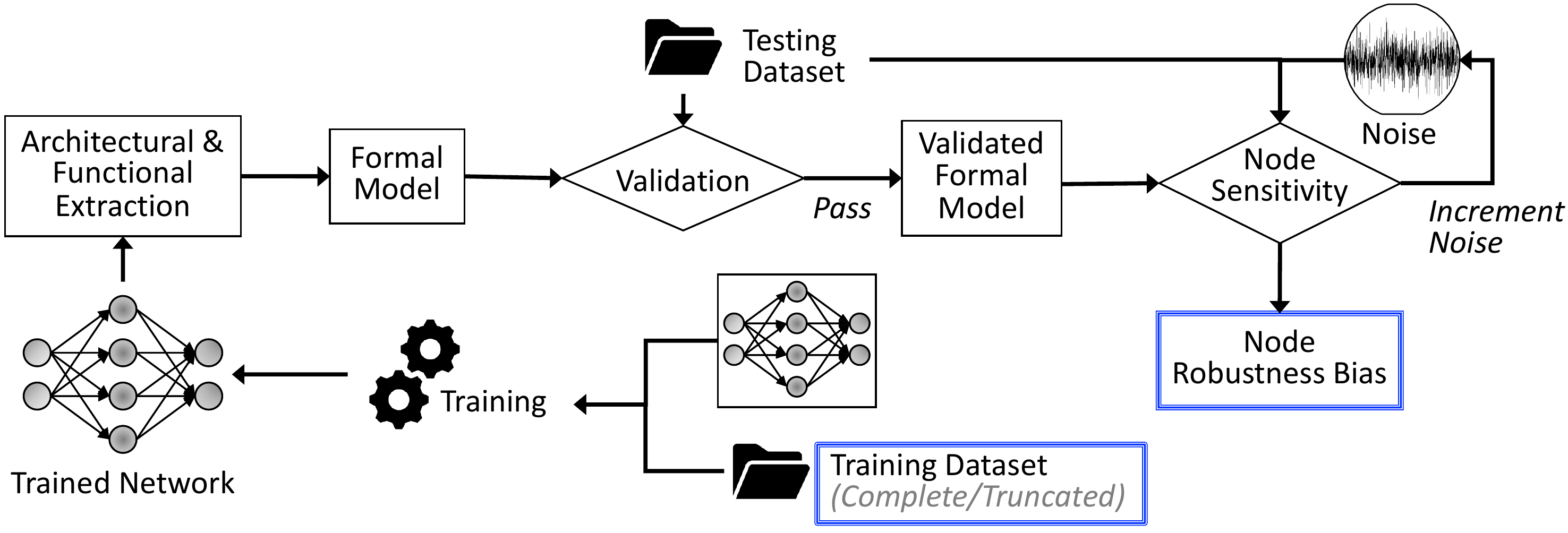}
\end{center}
\caption{Overview of our proposed framework for node (robustness) bias analysis. The training dataset could be complete or truncated to obtain a dataset to avoid long-tail distribution.}
\label{fig:meth} 
\end{figure}

Fig. \ref{fig:meth} provides an overview of our proposed analysis framework to study the node (robustness) bias of the trained networks. 
The architecture and parameter details of the trained DNN are initially used to construct the formal model of the network \cite{bhatti2022formal}. 
The formal model is validated using inputs from the testing dataset (i.e., the correct and model's computed output classification of the testing inputs are compared for consistency of results). 
The sensitivity of the input nodes is then analyzed using a probabilistic model checker as follows:
\begin{equation*}
    \mathds{P}_{=?}[\mathds{F}(\mathcal{N}(x) = \mathcal{N}(x\setminus x^i,x^i + \eta)) \land (\eta \leq N)],
\end{equation*}
    
where $i$ is the node under sensitivity analysis and $\mathds{F}(\mathcal{N}(x) = \mathcal{N}(x\setminus x^i,x^i + \eta))$ indicates that the network eventually provides correct output classification for input $x$. This is repeated iteratively, while gradually increasing the incident noise applied to the testing inputs. 
The exact node sensitivity results are then analyzed for individual input nodes to understand the node (robustness) bias. 
This is achieved initially using network trained on the complete dataset.

As indicated in earlier sections, the classification performance of the trained network may vary for networks trained on training datasets with long-tail distribution. 
Intuitively, this suggests that a network trained on dataset with an equal number of inputs from each class might address the discrepancy in classification performance across different classes \cite{class-imbalance-survey} and ensure that node (robustness) bias holds for the network. 
To test the hypothesis, we truncate the training dataset by deleting inputs from the head class(es), and repeat the analysis on this new dataset (which no longer has a long-tail distribution). 

\section{Case Study}
This section provides a case study to highlight the node (robustness) bias in a DNN trained on real dataset. This is followed by a discussion of the results and analysis.

\subsection{Experimental Setup}
We train single-hidden layer ReLU-based fully-connected neural networks on the top$-5$ relevant features of Leukemia dataset \cite{matlab-leukemia}. 
The training dataset comprises of $38$ inputs, with the head class (i.e., ALL Leukemia) constituting approximately $70\%$ of the dataset, while the tail class (i.e., AML Leukemia) constitutes the remaining dataset. 

The experiments were repeated $10$ times, while noting the results for the networks trained on the complete dataset. 
Similarly, $10$ networks were also trained on a truncated dataset via deleting randomly selected subset of inputs from the head class, before each training. 
This ensures an equal number of inputs from each class. 
Storm model checker was used for the quantitative verification of node sensitivity. All experiments were run on AMDRyzen Threadripper $2990WX$ processors running Ubuntu $18.04$ LTS operating system.

\subsection{Results and Discussion}

\begin{figure}[ht]
\begin{center}
\includegraphics[width=\linewidth]{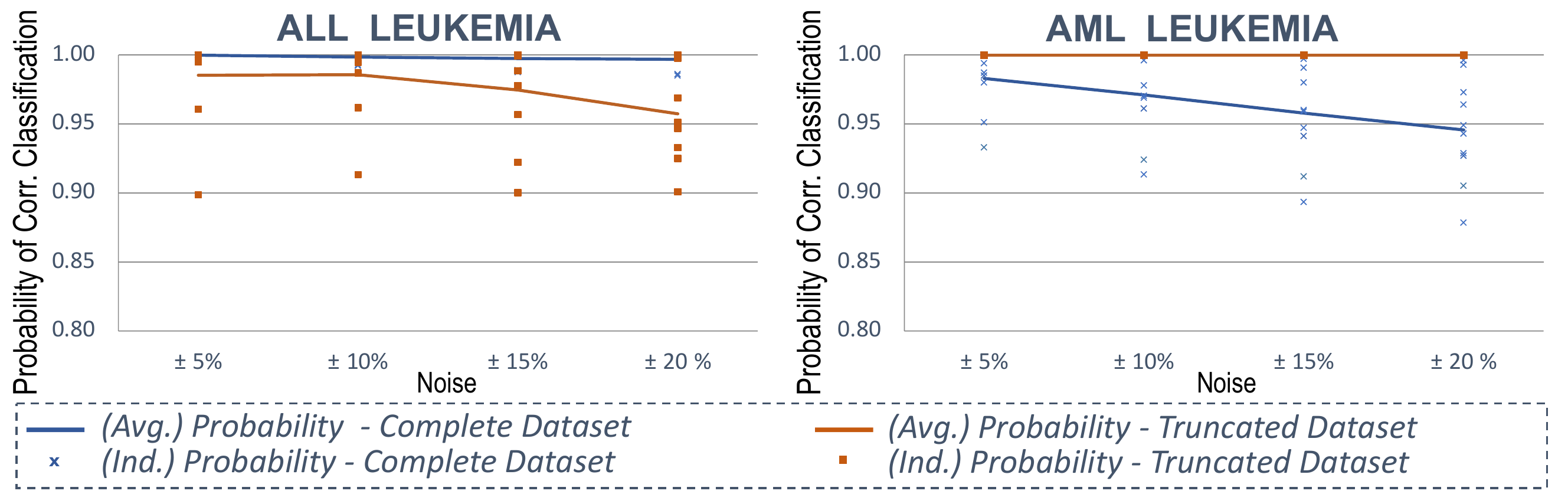}
\end{center}
\caption{Networks trained on long-tail distribution indicate robustness bias for AML (i.e., tail class). Truncation of training dataset prior to training towards ALL, but only for large noise.}
\label{fig:bias} 
\end{figure}

As indicated earlier, the training dataset of the Leukemia dataset composes a long-tail distribution. 
Hence, the networks trained on it delineate robustness bias, with the increase in incident noise gradually decreasing the probability of correct classification of AML, but not for ALL. This is presented by the blue lines in Fig. \ref{fig:bias}. 
The truncation of ALL inputs from the training dataset, in turn, generates networks that appear unbiased for at low incident noise (see the orange lines in Fig. \ref{fig:bias}). However, for large noise, the classification probability of ALL starts to decrease whereas the AML is correctly classified with a probability of $\sim 1.0$. 
This suggests that long-tail distribution is only a component of a much more complicated problem, leading to robustness bias. 
Hence, while avoiding long-tail distribution addresses robustness bias for smaller noise, the strategy alone may not be sufficient bias reduction strategy for inputs exposed to larger noise. 

\begin{figure*}[ht]
\begin{center}
\includegraphics[width=\linewidth]{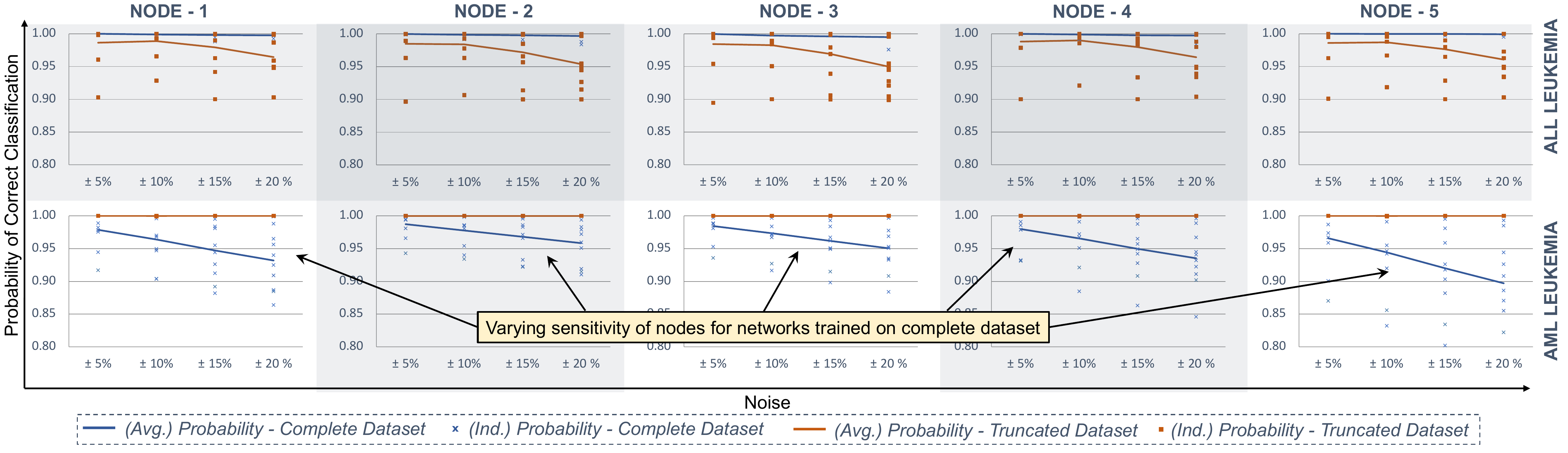}
\end{center}
\caption{The node sensitivity against negative noise for ALL and AML Leukemia. The lines present the average probability results while points present result from each (of the $10$) experiments. The sensitivity varies for networks trained on complete and truncated datasets.}
\label{fig:sens_neg} 
\end{figure*}

Similar trends are observed from the results of node sensitivity for negative noise, as shown in Fig. \ref{fig:sens_neg}. 
Truncation of training dataset leads to network's input nodes having approximately equal classification probability for small noise. However, the probability of correct classification gradually decreases for ALL at higher noise. 
It can also be observed for the networks trained on the original dataset that the sensitivity of different nodes is visibly different, as observed by the corresponding gradients of the blue lines for AML.

\begin{figure*}[ht]
\begin{center}
\includegraphics[width=\linewidth]{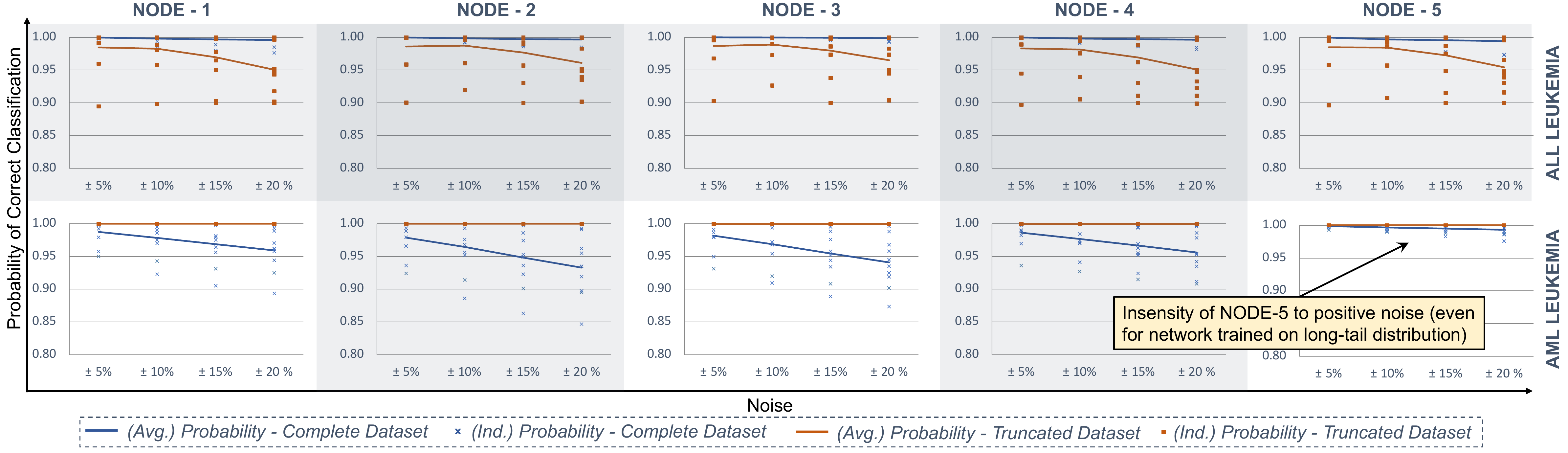}
\end{center}
\caption{The node sensitivity against positive noise for ALL and AML Leukemia. The lines present the average probability results while points present result from each (of the $10$) experiments. The sensitivity varies for networks trained on complete and truncated datasets.}
\label{fig:sens_pos} 
\end{figure*}

The observations for the analysis with positive incident noise provide similar results in case of ALL (see Fig. \ref{fig:sens_pos}). 
However, a stark difference can be observed for the sensitivity results of NODE$-5$, for AML. 
Where with negative noise, NODE$-5$ was observed to be most sensitive to noise, in case of positive noise, the node is observed to be very insensitive. 
This indicates a significant node (robustness) bias for networks trained on the original dataset. 
It is also interesting to note that the similar node (robustness) bias is not observed for the same input node for inputs belonging to ALL Leukemia, for trained on both the original and truncated datasets. 


Such sensitivity of NODE$-5$ suggests a biased learning of the node for AML. It can also be observed through Table \ref{tab:stats} that NODE$-5$ has the largest variance among all nodes, for AML. This could be a possible reason for the strange sensitivity of the node, and the subsequent node (robustness) bias. However, given a similar condition for NODE$-3$ for ALL (i.e., having the largest variance for ALL), a similar discrepancy for node (robustness) bias, between positive and negative noise, is not observed. 
\begin{table}[ht]
\caption{Variance of input node values in training datasets. The extrema of the variances, for each class, are {\ul underlined}}
\label{tab:stats}
\begin{tabular}{lrr}
\multicolumn{3}{c}{\textbf{Training Dataset}}       \\ \toprule
         & ALL Variance (x$10^3$) & AML Variance (x$10^3$)  \\ \midrule
NODE - 1 & $114.27$               & $129.72$                                      \\
NODE - 2  & $81.21$                & {  {\ul $11.71$}}          \\
NODE - 3 & {\ul $5531.62$}        & $231.77$                                      \\
NODE - 4 & {\ul $45.24$}          & $284.02$                                      \\
NODE - 5 & $156.40$               & {\ul $2271.00$}  \\ \bottomrule 
\end{tabular}
\end{table}


\section{Related Works}
Long-tail distribution is a widely studied challenge in DNN research community, since it is often associated with a varying classification performance of the network for \textit{head} and \textit{tail} classes \cite{class-imbalance-survey,2022longtail}. 
Numerous pre-training, training, fine-tuning and transfer-learning approaches have been proposed to ensure the overall classification performance stays consistent for all classes \cite{zhang2021bag,longtail_lossfunc,trans_learning,chawla2002smote,feature_trans_learn,tail-noise}.
Recently, robustness bias was shown to be a concern for DNNs with noisy inputs \cite{rbias}. 
It was also shown that such a bias may inadvertently be a consequence of long-tail distribution of the training dataset \cite{bhatti2022formal}. 
It was also shown that the such bias is not just a result of long-tail distribution, but may also be the consequence of the unequal data representation of the training data samples \cite{naseer2023unbiasednets}.  
However, the robustness bias of the network was studied for the inputs as a whole, ignoring the bias for individual input nodes. 

Orthogonally, the sensitivity of input nodes has also been explored in the literature \cite{feature_bias,yao2003sensitivity}. 
These works aim to identify the most relevant input features for learning, the determination of which, could then be used for network simplification (for instance, via input pruning \cite{input-sens1,chen2020sensitivity}) or ensuring the privacy of sensitive input features \cite{sens4fair1_dnn,sens4fair2_dnn}.
However, this is essentially different from our work, which caters for the sensitivity of input nodes under the impact of noise, which could be different from that on clean inputs. 


\section{Conclusion and Future Work}

Past years have not only seen a rise in the use of deep neural networks (DNNs) in real-world applications, but also an awareness of the vulnerabilities of these DNNs leading to their undesirable performance. 
Among the concerns arising regarding DNNs is the inconsistent classification of these networks across the output classes, often resulting from long-tail distribution of the training dataset. 
Existing literature already highlights the (robustness) bias as a possible consequence of such distributions. 
However, in this work, we shed light on the bias of the DNNs beyond simply the varying classification performance across different output classes. 
To the best of our knowledge, this is the first work exposing the varying bias of input nodes, for the different output classes, for DNNs trained on long-tail distribution data.

Through our proposed framework and case study, we also explore a possible link between variance of node values in the training dataset and the impact of removing samples of head class(es) of the distribution during training. 
However, the exact cause of such node bias is still up for debate, and requires further analysis with larger datasets and attention from the research community.

\bibliographystyle{IEEEtran}
\bibliography{ref}

\begin{thebibliography}{10}
\providecommand{\url}[1]{#1}
\csname url@samestyle\endcsname
\providecommand{\newblock}{\relax}
\providecommand{\bibinfo}[2]{#2}
\providecommand{\BIBentrySTDinterwordspacing}{\spaceskip=0pt\relax}
\providecommand{\BIBentryALTinterwordstretchfactor}{4}
\providecommand{\BIBentryALTinterwordspacing}{\spaceskip=\fontdimen2\font plus
\BIBentryALTinterwordstretchfactor\fontdimen3\font minus
  \fontdimen4\font\relax}
\providecommand{\BIBforeignlanguage}[2]{{%
\expandafter\ifx\csname l@#1\endcsname\relax
\typeout{** WARNING: IEEEtran.bst: No hyphenation pattern has been}%
\typeout{** loaded for the language `#1'. Using the pattern for}%
\typeout{** the default language instead.}%
\else
\language=\csname l@#1\endcsname
\fi
#2}}
\providecommand{\BIBdecl}{\relax}
\BIBdecl

\bibitem{health}
A.~Esteva~et al., ``A guide to deep learning in healthcare,''
  \emph{\textbf{{Nat. Med.}}}, vol.~25, no.~1, pp. 24--29, 2019.

\bibitem{auto_drive}
G.~Li~et al., ``A deep learning based image enhancement approach for autonomous
  driving at night,'' \emph{\textbf{{KBS}}}, vol. 213, p. 106617, 2021.

\bibitem{2022longtail}
Y.~Fu~et al., ``Long-tailed visual recognition with deep models: A
  methodological survey and evaluation,'' \emph{\textbf{Neurocomputing}}, 2022.

\bibitem{arythmia-dataset}
G.~B. Moody~et al., ``The impact of the mit-bih arrhythmia database,''
  \emph{\textbf{{Eng. Med. Biol. Mag.}}}, vol.~20, no.~3, pp. 45--50, 2001.

\bibitem{imdb-dataset}
R.~R. et~al., ``Deep expectation of real and apparent age from a single image
  without facial landmarks,'' \emph{\textbf{{IJCV}}}, vol. 126, no. 2-4, pp.
  144--157, 2018.

\bibitem{wu2014wafer}
M.-J. Wu~et al., ``Wafer map failure pattern recognition and similarity ranking
  for large-scale data sets,'' \emph{\textbf{{Trans. Semicond. Manuf.}}},
  vol.~28, no.~1, pp. 1--12, 2014.

\bibitem{bhatti2022formal}
I.~T. Bhatti~et al., ``A formal approach to identifying the impact of noise on
  neural networks,'' \emph{\textbf{{Commun. ACM}}}, vol.~65, no.~11, pp.
  70--73, 2022.

\bibitem{input-sens1}
J.~M. Zurada~et al., ``Sensitivity analysis for minimization of input data
  dimension for feedforward neural network,'' in \emph{\textbf{{ISCAS}}},
  vol.~6.\hskip 1em plus 0.5em minus 0.4em\relax IEEE, 1994, pp. 447--450.

\bibitem{chen2020sensitivity}
S.~Chen~et al., ``Sensitivity analysis to reduce duplicated features in ann
  training for district heat demand prediction,'' \emph{\textbf{{Energy \&
  AI}}}, vol.~2, p. 100028, 2020.

\bibitem{sens4fair1_dnn}
Z.~Zhang~et al., ``Hate speech detection: A solved problem? the challenging
  case of long tail on twitter,'' \emph{\textbf{{Semantic Web}}}, vol.~10,
  no.~5, pp. 925--945, 2019.

\bibitem{sens4fair2_dnn}
A.~Morales~et al., ``Sensitivenets: Learning agnostic representations with
  application to face images,'' \emph{\textbf{{TPAMI}}}, vol.~43, no.~6, pp.
  2158--2164, 2020.

\bibitem{class-imbalance-survey}
J.~L. Leevy~et al., ``A survey on addressing high-class imbalance in big
  data,'' \emph{\textbf{{J. Big Data}}}, vol.~5, no.~1, pp. 1--30, 2018.

\bibitem{matlab-leukemia}
S.~Khan~et al., ``A novel fractional gradient-based learning algorithm for
  recurrent neural networks,'' \emph{\textbf{{CSSP}}}, vol.~37, no.~2, pp.
  593--612, 2018.

\bibitem{zhang2021bag}
Y.~Zhang~et al., ``Bag of tricks for long-tailed visual recognition with deep
  convolutional neural networks,'' in \emph{\textbf{Proc. {AAAI}}}, vol.~35,
  no.~4, 2021, pp. 3447--3455.

\bibitem{longtail_lossfunc}
D.~Samuel~et al., ``Distributional robustness loss for long-tail learning,'' in
  \emph{\textbf{Proc. {ICCV}}}, 2021, pp. 9495--9504.

\bibitem{trans_learning}
Y.-X. Wang~et al., ``Learning to model the tail,'' in \emph{\textbf{Proc.
  {NeurIPS}}}, e.~a. I.~Guyon, Ed., vol.~30.\hskip 1em plus 0.5em minus
  0.4em\relax Curran Associates, Inc., 2017.

\bibitem{chawla2002smote}
N.~V. Chawla~et al., ``Smote: synthetic minority over-sampling technique,''
  \emph{\textbf{{JAIR}}}, vol.~16, pp. 321--357, 2002.

\bibitem{feature_trans_learn}
X.~Yin~et al., ``Feature transfer learning for face recognition with
  under-represented data,'' in \emph{\textbf{Proc. {CVPR}}}, 2019, pp.
  5704--5713.

\bibitem{tail-noise}
J.~Liu~et al., ``Deep representation learning on long-tailed data: A learnable
  embedding augmentation perspective,'' in \emph{\textbf{Proc. {CVPR}}}, 2020,
  pp. 2970--2979.

\bibitem{rbias}
V.~Nanda~et al., ``{Fairness Through Robustness: Investigating Robustness
  Disparity in Deep Learning},'' in \emph{\textbf{Proc. {FAccT}}}, 2021, pp.
  466--477.

\bibitem{naseer2023unbiasednets}
M.~Naseer~et al., ``{UnbiasedNets}: a dataset diversification framework for
  robustness bias alleviation in neural networks,'' \emph{\textbf{ML}}, pp.
  1--28, 2023.

\bibitem{feature_bias}
D.~Dai~et al., ``Rethinking the image feature biases exhibited by deep
  convolutional neural network models in image recognition,''
  \emph{\textbf{{CAAI Trans. Intell. Technol.}}}, 2022.

\bibitem{yao2003sensitivity}
J.~Yao, ``Sensitivity analysis for data mining,'' in \emph{\textbf{Proc.
  {NAFIPS}}}.\hskip 1em plus 0.5em minus 0.4em\relax IEEE, 2003, pp. 272--277.

\end{thebibliography}

\end{document}